\documentclass[letterpaper]{article} 
\usepackage{aaai2026}  
\usepackage{times}  
\usepackage{helvet}  
\usepackage{courier}  
\usepackage[hyphens]{url}  
\usepackage{graphicx} 
\usepackage{natbib}  
\usepackage{caption} 
\usepackage{algorithm}
\usepackage{algorithmic}

\usepackage{newfloat}
\usepackage{listings}
\DeclareCaptionStyle{ruled}{labelfont=normalfont,labelsep=colon,strut=off} 
\floatstyle{ruled}
\newfloat{listing}{tb}{lst}{}
\floatname{listing}{Listing}
\usepackage{xcolor}         
\usepackage{amsmath}
\usepackage{amssymb}
\usepackage{mathtools}
\usepackage{amsthm}
\usepackage{microtype}
\usepackage{graphicx}
\usepackage{subcaption}
\usepackage{caption}
\usepackage{xcolor}
\usepackage{booktabs} 
\usepackage{mwe,tikz}
\usepackage[percent]{overpic}
\usepackage{subcaption}
\usepackage{makecell}
\usepackage{multirow} 
\usepackage{dsfont}
\DeclareCaptionStyle{ruled}{labelfont=normalfont,labelsep=colon,strut=off} 
\floatstyle{ruled}
\newfloat{listing}{tb}{lst}{}
\floatname{listing}{Listing}
\title{CalibrateMix: Guided-Mixup Calibration of Image Semi-Supervised Models}
\author{
    Mehrab Mustafy Rahman \mbox{    }\mbox{    }\mbox{    }\mbox{    }
    Jayanth Mohan \mbox{    }\mbox{    }\mbox{    }\mbox{    }
    Tiberiu Sosea \mbox{    }\mbox{    }\mbox{    }\mbox{    }
    Cornelia Caragea
}

\affiliations{
Computer Science \\
    University of Illinois Chicago\\
    \vspace{1mm}
    \color{blue}{\texttt{mrahm@uic.edu}\mbox{    }\mbox{    }\mbox{    }\mbox{    }
    \texttt{jmoha11@uic.edu}\mbox{    }\mbox{    }\mbox{    }\mbox{    }
    \texttt{tsosea2@uic.edu}\mbox{    }\mbox{    }\mbox{    }\mbox{    }
    \texttt{cornelia@uic.edu}\mbox{    }\mbox{    }\mbox{    }\mbox{    }}
}

\begin{document}

\maketitle

\begin{abstract}
Semi-supervised learning (SSL) has demonstrated high performance in image classification tasks by effectively utilizing both labeled and unlabeled data. However, existing SSL methods often suffer from poor calibration, with models yielding overconfident predictions that misrepresent actual prediction likelihoods. Recently, neural networks trained with {\tt mixup} that linearly interpolates random examples from the training set have shown better calibration in supervised settings. However, calibration of neural models remains under-explored in semi-supervised settings. Although effective in supervised model calibration, random mixup of pseudolabels in SSL presents challenges due to the overconfidence and unreliability of pseudolabels. In this work, we introduce CalibrateMix, a targeted mixup-based approach that aims to improve the calibration of SSL models while maintaining or even improving their classification accuracy. Our method leverages training dynamics of labeled and unlabeled samples to identify ``easy-to-learn'' and ``hard-to-learn'' samples, which in turn are utilized in a targeted mixup of easy and hard samples. Experimental results across several benchmark image datasets show that our method achieves lower expected calibration error (ECE) and superior accuracy compared to existing SSL approaches.
\end{abstract}

\begin{links}
    \link{Code}{https://github.com/mehrab-mustafy/CalibrateMix}
\end{links}

\section{Introduction}
\label{sec:intro}

Deep neural networks (DNNs) \cite{lecun2015deep, mathew2021deep} have achieved remarkable success across a wide range of computer vision tasks, including image classification \cite{lu2007survey}, object detection \cite{papageorgiou1998general}, and semantic segmentation \cite{minaee2021image}. However, alongside accuracy, the predictive confidence of the models plays a vital role in real-world decision-making applications. For example, in critical applications such as autonomous driving, medical diagnosis, and disaster response, models must be accurate as well as reliably indicate when they are uncertain, so that additional safety measures can be triggered. For this reason, quantifying predictive uncertainty and calibration of the DNNs is a pivotal component toward building more reliable models. Despite strong performance, DNNs often suffer from poor calibration \cite{guo2017calibration}, which means that the predictive confidence likely overestimates the model's true accuracy. A key reason behind this is that modern DNNs are trained using one-hot encoded labels and the cross-entropy loss, which assumes that every training sample belongs with full certainty to a single class. This forces the model to assign the entire probability mass to a single class label, which in turn suppresses any expression of uncertainty even for ``ambiguous'' samples. To overcome the issues of overconfidence, label smoothing \cite{muller2019does} has been introduced. By softening the target distribution during training, label smoothing regularizes the model's output probabilities, encouraging it to remain uncertain where appropriate. More recently, \citet{thulasidasan2019mixup} explored the use of mixup training \cite{ZhangCDL18mixup} for improving model calibration, which creates augmented samples through convex combinations of input samples and their labels. Mixup distributes its probability into two classes, which introduces entropy, prevents overconfidence, and has proven to be an effective tool in model calibration.

\begin{figure*}[h]
    \centering
    \begin{subfigure}[b]{0.24\textwidth}
        \centering
        \includegraphics[width=\textwidth]{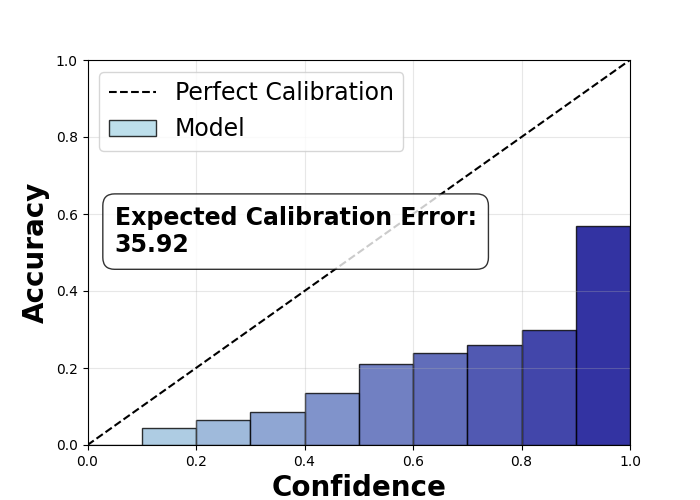}
        \caption{}
        \label{subfig3}
    \end{subfigure}
    \hfill
    \begin{subfigure}[b]{0.24\textwidth}
        \centering
        \includegraphics[width=\textwidth]{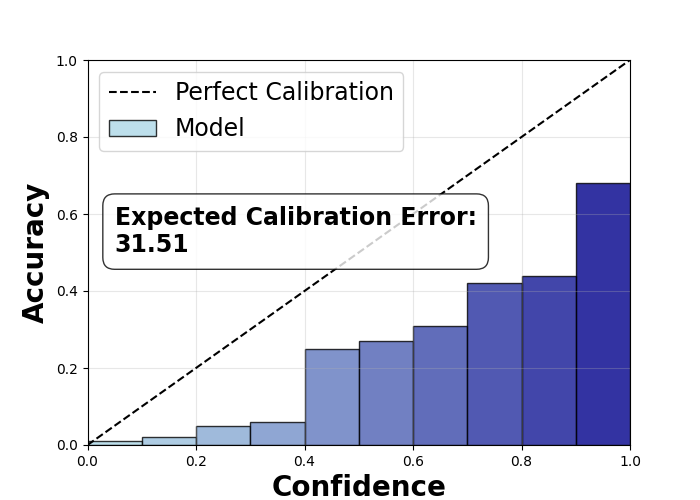}
        \caption{}
        \label{subfig4}
    \end{subfigure}
    \begin{subfigure}[b]{0.24\textwidth}
        \centering
        \includegraphics[width=\textwidth]{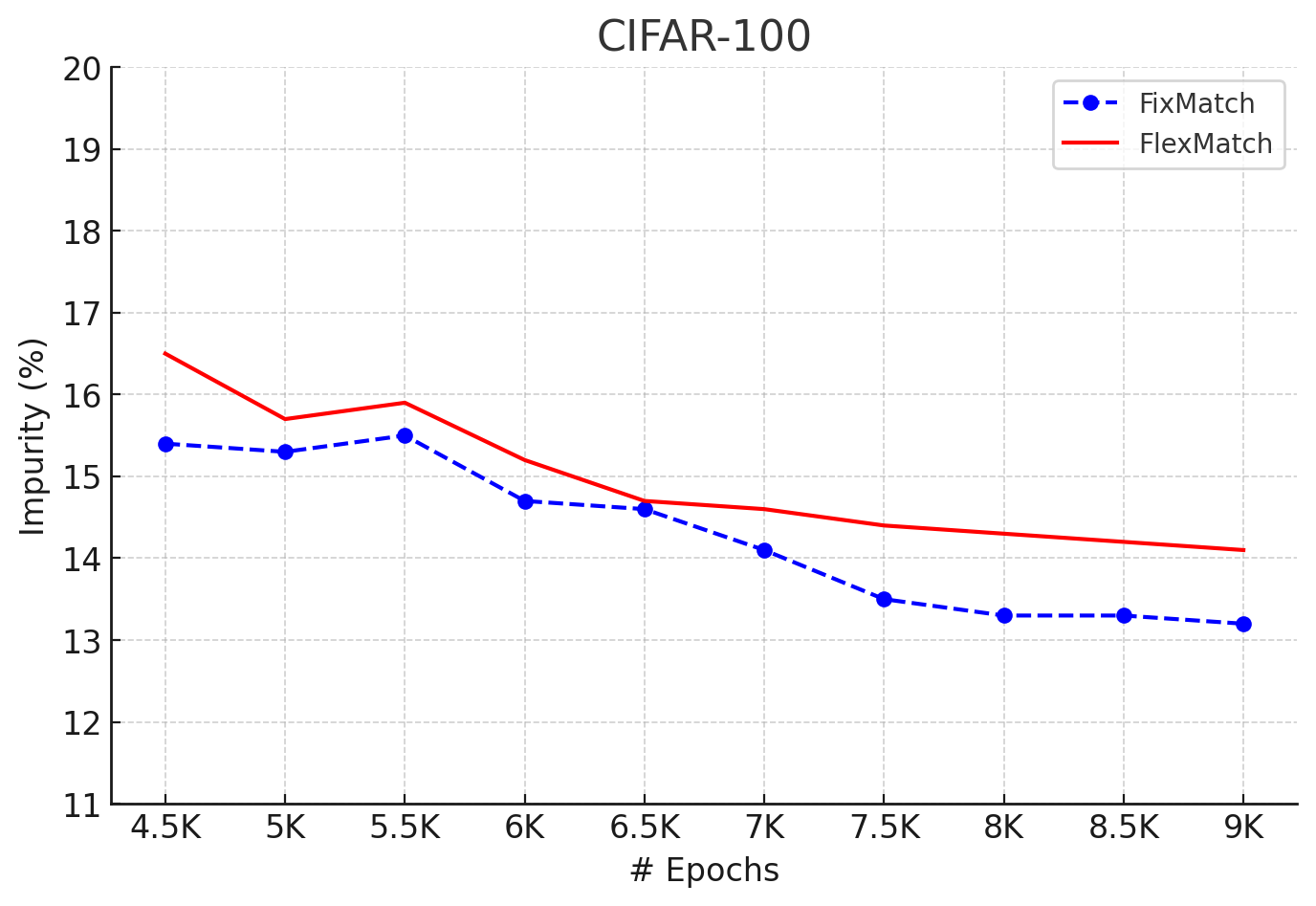}
        \caption{}
        \label{subfig2}
    \end{subfigure}
    \hfill
    \begin{subfigure}[b]{0.24\textwidth}
        \centering
        \includegraphics[width=\textwidth]{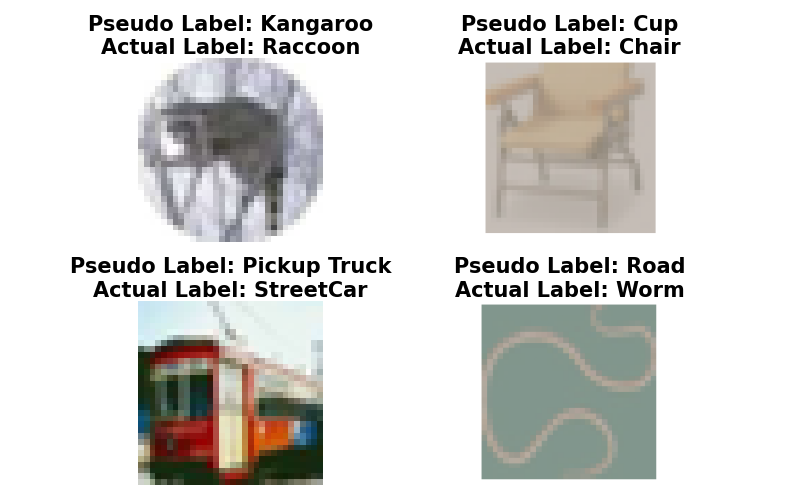}
        \caption{}
        \label{subfig1}
    \end{subfigure}
    \hfill

    \caption{
    (a) FixMatch reliability diagram on CIFAR-100; (b) FlexMatch reliability diagram on CIFAR-100; (c) Impurity comparison on CIFAR-100; (d) Incorrect pseudo-labels at end of training for FixMatch(right) and FlexMatch(left).
    }
    \label{fig:fig1}
\end{figure*}

However, this approach primarily targets the fully supervised setting, which requires a large amount of labeled data. With the emergence of AI in all domains, it is impractical to obtain large amounts of annotated data for every domain. To solve this issue, Semi-supervised learning (SSL) \cite{chapelle2009semi} can be an effective strategy to leverage large amounts of unlabeled examples during training to boost model performance. Despite mixup being successful in supervised learning, mixing up unlabeled examples in an SSL setting poses certain challenges due to the uncertainty of pseudo-label correctness, especially at the early iterations of training. This makes it critical to ensure the quality of pseudo-labels before incorporating them into the learning process. To ensure the quality of pseudo-labels, a popular SSL method to learn from unlabeled examples is pseudo-labeling \cite{lee2013pseudo}, which leverages a model to make predictions on unlabeled examples and assign them pseudo-labels, which are in turn used as (pseudo) ground truth during training. To ensure that correct pseudo-labels are used for model training, modern SSL frameworks such as FixMatch \cite{sohn2020fixmatch} and FlexMatch \cite{zhang2021flexmatch} utilize high confidence thresholds to maintain data quality and filter out potentially incorrect examples. However, the calibration of these SSL models is not well studied, and we found empirical evidence that SSL models also suffer from poor calibration as shown in the reliability diagrams of FixMatch and FlexMatch on CIFAR-100 in Figures \ref{subfig3} and \ref{subfig4}, respectively. The diagrams, which plot accuracy as a function of confidence, show that the confidence estimates of the models are not indicative of their correctness. Notably, FixMatch and FlexMatch predictions with confidences higher than $90\%$ have less than $65\%$ accuracy, contradicting the assumption that employing high confidence thresholds leads to high pseudo-label quality.

As shown in Figure \ref{subfig2}, the impurity of unlabeled data for both FixMatch and FlexMatch on CIFAR-100 \cite{krizhevsky2009learning} is higher than $13\%$, indicating that more than $13\%$ of the unlabeled data is utilized with incorrect pseudo-labels during training even at the later stages of training at 9000 epoch. Figure \ref{subfig1} shows examples of incorrect pseudo-labels introduced by both FixMatch and FlexMatch at the end of training. These incorrect pseudo-labels arise due to miscalibrated model predictions, manifested here as overconfidence in potentially incorrect predictions. This calibration gap in SSL can not be fully solved using random mixup. When incorrect pseudo-labels are used in mixup, the interpolation process can propagate label noise across training samples. This not only reinforces errors but also makes them harder to correct. For deep learning models that are highly over-parameterized and capable of achieving near-zero training error, such reinforcement of incorrect labels can lead to severe overfitting to mislabeled data. Thus, without addressing the underlying calibration issue and pseudo-label reliability, mixup can become a mechanism for error amplification rather than a regularization strategy. Hence, we investigate the calibration of SSL models under mixup and propose an enhanced mixup strategy.

In this paper, we propose CalibrateMix, a novel, targeted mixup-based \cite{ZhangCDL18mixup} framework that enhances the confidence calibration of SSL models. CalibrateMix monitors the training dynamics of unlabeled samples during training by keeping track of margins \cite{bartlett2017spectrally,elsayed2018large,sosea2023marginmatch} of the outputs of the model. These margins are utilized to characterize each sample based on learning difficulty into two categories:``easy-to-learn'' samples, which the model perceives to be correctly pseudo-labeled and ``hard-to-learn'' samples, which the model perceives to be potentially incorrect or ambiguous. Then, our method performs targeted mixup, combining easy-to-learn labeled samples with hard-to-learn unlabeled samples, as well as hard-to-learn labeled samples with easy-to-learn unlabeled samples. This mixup strategy has two main advantages: First, the presence of easy-to-learn samples ensures that the resulting pseudo-label of the mixed sample is qualitative, while the presence of the hard-to-learn samples ensures a more difficult environment for the model to learn from, which was shown to benefit SSL training \cite{xie2020self}. Second, CalibrateMix is easy to use in practice and can be applied on top of the most popular SSL frameworks, such as FixMatch or FlexMatch.

To showcase the benefits of our method, we carry out extensive experiments on well-established small-scale SSL setups - CIFAR10, CIFAR100 \cite{krizhevsky2009learning}, SVHN \cite{netzer2011reading}, STL10 \cite{pmlr-v15-coates11a}; and on two large-scale benchmarks - ImageNet \cite{deng2009imagenet} and WebVision \cite{li2017webvision}, where we show that CalibrateMix outperforms strong baselines and previous works both in terms of accuracy and expected calibration error (ECE).
Notably, CalibrateMix boosts the performance of FixMatch by $8.51\%$ in ECE and by $1.54\%$ in error rate on CIFAR-100 using only 25 labels per class. On the larger datasets, e.g., WebVision, we also see an improvement of $2\%$ ECE on FixMatch and FlexMatch without loss in error rates.

Our contributions are as follows: 
\begin{enumerate}
    \item We introduce CalibrateMix, a novel targeted mixup framework that is primarily designed to improve the confidence calibration of semi-supervised learning (SSL) models. 
    \item Through extensive experiments on standard benchmark image datasets, we demonstrate that CalibrateMix outperforms existing methods in terms of Expected Calibration Error (ECE), while also often achieving lower error rates compared to multiple state-of-the-art SSL setups. Our method is compatible with any SSL frameworks, and we demonstrate its effectiveness by improving both calibration and, in multiple cases, accuracy when integrated with FixMatch, FlexMatch, and SoftMatch.
    \item We carry out extensive ablation studies to understand the contribution of various components to the success of CalibrateMix, as well as an error analysis to analyze the calibration and correctness of predictions in practice.
\end{enumerate}

\section{Related Work}
\label{sec:relwork}

Despite achieving high accuracy, modern deep networks frequently remain miscalibrated, producing overly confident predictions \cite{guo2017calibration,minderer2021revisiting}. Several effective methods have been proposed in supervised learning settings to mitigate calibration issues proactively. Dropout, initially introduced as a regularization technique, doubles as a Bayesian uncertainty estimator at inference via Monte Carlo sampling, providing improved calibration by averaging predictions across multiple stochastic forward passes \cite{srivastava2014dropout}. Another well established approach is label smoothing, which softens the one-hot encoded targets, thereby discouraging extreme predictions and significantly enhancing calibration and generalization performance, especially under limited or imbalanced data scenarios \cite{muller2019does}.  Mixup training, another promising approach, interpolates pairs of training samples and their labels, generating augmented data points that promote smoother predictive boundaries. Thulasidasan et al. \cite{thulasidasan2019mixup} explicitly verified mixup's effectiveness in reducing expected calibration error (ECE), highlighting its capability in generating well-calibrated confidence scores and robust predictive uncertainty. 
Further developments, such as RegMixup, have validated that Mixup based regularization not only enhances calibration but also robustness under distribution shifts, particularly beneficial in out-of-distribution (OOD) detection tasks \cite{pinto2022using}. 

Other studies demonstrated that employing focal loss, which emphasizes learning from challenging predictions, also effectively mitigates model overconfidence, enhancing calibration and model reliability \cite{mukhoti2020calibrating}. While supervised calibration methods have matured significantly, translating these advancements directly to semi-supervised learning (SSL) remains challenging. SSL leverages limited labeled data alongside abundant unlabeled data to enhance model learning. Prominent SSL frameworks, such as FixMatch \cite{sohn2020fixmatch}, FlexMatch \cite{zhang2021flexmatch}, and SoftMatch \cite{chen2023softmatch}, primarily rely on pseudo-labeling and consistency regularization. FixMatch employs fixed high confidence thresholds to filter pseudo-labels, improving overall accuracy but inadvertently exacerbating calibration issues by neglecting calibration aware decision-making. Similarly, FlexMatch introduces a dynamic curriculum based threshold adjustment, enhancing class specific learning but still neglecting explicit calibration mechanisms \cite{zhang2021flexmatch}. SoftMatch further improves pseudo-label quality through adaptive weighting strategies; however, calibration remains an implicit, secondary consideration rather than a deliberate design goal \cite{chen2023softmatch}. Consequently, SSL models often remain miscalibrated, assigning overly confident pseudo-labels to unlabeled samples, which compromises their accuracy and robustness. Recent SSL approaches, such as MarginMatch \cite{sosea2023marginmatch}, SequenceMatch \cite{nguyen2024sequencematch}, and FineSSL \cite{gan2024erasing}, indirectly target pseudo-label reliability and robustness through pseudo-margin analysis, consistency across augmentations, and robust finetuning strategies, respectively. While these techniques implicitly reduce calibration errors, their main emphasis remains on enhancing pseudo-label quality or robustness rather than directly addressing calibration gaps. Thus, explicit calibration remains largely unaddressed. Motivated by this necessity, we propose CalibrateMix, a targeted mixup-based SSL approach explicitly designed to improve both model calibration and predictive accuracy.

\section{CalibrateMix}

\label{sec:approach}

\textbf{Notation} 
Let $D_{L}$ = $\{(x_1, y_1),\cdots,(x_{B_{L}}, y_{B_{L}}) \}$ be a batch of labeled samples of size $B_{L}$ and $D_{U}$ = $\{\hat{x}_1,\cdots,\hat{x}_{B_{U}} \}$ be a batch of unlabeled samples of size $B_{U}$. 

\subsection{Background}

 Most deep neural networks (DNNs) trained for classification are trained using one-hot encoded labels, where the entire probability mass is assigned to a single class. This leaves no room for uncertainty during training. As a result, models tend to become overconfident in their predictions. Hence, it is not surprising that modern DNNs are poorly calibrated \cite{guo2017calibration}. To prevent this overconfidence, \citet{thulasidasan2019mixup} explored the impacts of mixup training in supervised settings and found that mixup has proven to be effective in reducing miscalibration in supervised settings. Mixup training creates vicinity samples for training, effectively creating more samples for the model to learn from. The mixup augmented samples are generated by the following rule as mentioned in \citet{ZhangCDL18mixup}:
\begin{equation}
\tilde{x} = \gamma x_i + (1 - \gamma) x_j
\end{equation}
\begin{equation}
\tilde{y} = \gamma y_i + (1 - \gamma) y_j
\end{equation}
where $x_i$ and $x_j$ are any two randomly selected samples from the train set and $y_i$ and $y_j$ are their corresponding one-hot labels. Mixup-augmented sample \( \tilde{x} \) is generated by interpolating between \( x_i \) and \( x_j \). Similarly, the corresponding label \( \tilde{y} \) is obtained by mixing \( y_i \) and \( y_j \). The mixup coefficient \( \gamma \) comes from the Beta distribution, with the hyperparameter \( \alpha \) controlling the level of interpolation between the two samples. Unlike standard one-hot labels, when $y_i
\neq y_j$, the mixup label \( \tilde{y} \) distributes the probability mass across two classes rather than a single class. This distribution introduces entropy and uncertainty into the training process, preventing the model from becoming overly confident in its predictions.

In SSL, selecting samples at random for mixup can be harmful because of the unreliability of pseudo-labels, especially in the early stages of training when models are more prone to errors, and because of the confirmation bias. This is particularly severe for difficult samples. From Figure~\ref{fig:fig1}, we observe that the pseudo-labels may remain incorrect at the end of training, and SSL models can produce high-confidence incorrect predictions. Such miscalibrated predictions not only hamper model learning progress but also risk propagating errors across training iterations. Given that pseudo-labeled data typically outnumbers labeled data in SSL settings, random pairing during mixup increases the chance of combining two erroneous pseudo-labels. This can yield misleading targets that further reinforce incorrect representations. Hence, to reduce the impact of noisy labels in training and ensure quality, it is important to avoid mixup between two difficult samples or two pseudo-labeled samples.

\subsection{Proposed Approach: CalibrateMix}
To address the limitations of standard random mixup in SSL and to deal with the propagation of overconfident or noisy pseudo-labels we propose \textbf{CalibrateMix}, a targeted mixup-based framework. CalibrateMix performs a controlled mixup between labeled and pseudo-labeled samples guided by their learning difficulty to generate higher-quality augmented training samples. The inclusion of labeled samples in the mixup ensures that no two pseudo-labeled samples take part in the mixup at the same time, while the inclusion of pseudo-labeled samples introduces entropy and uncertainty. This helps prevent the reinforcement of noise and encourages the model to express necessary uncertainty, leading to better calibrated confidence estimates. Hence, the core advantage of CalibrateMix lies in its structured pairing strategy that mixes labeled and unlabeled samples based on their learning difficulty. To quantify the learning difficulty of the samples of the labeled data we monitor the Area Under the Margin (AUM) \cite{pleiss2020identifying} of the outputs of the model at each iteration. For the unlabeled data, we monitor the Average Pseudo Margin (APM) \cite{sosea2023marginmatch} of the outputs of the model at each iteration. We do this because relying solely on the model’s current prediction confidence is insufficient. Confidence at a single iteration does not reliably reflect the sample’s margin or the correctness of its pseudo-label, as shown in \citet{sosea2023marginmatch}. To calculate APM for an unlabeled sample $\hat{x}_i$, we use pseudo-margins defined as:
\begin{equation}
\text{PM}^t_{c}(\hat{x}_i) = z_{c}(\hat{x}_i) - \max_{j \neq c} z_j(\hat{x}_i)
\end{equation}
where at iteration $t$, $z_{c}(\hat{x}_i)$ is the logit corresponding to sample $\hat{x}_i$ for assigned pseudo-label $c$ and $\max_{j \neq c} z_j(\hat{x}_i)$ is the largest other logit corresponding to a label $j$ other than $c$ for the same sample $\hat{x}_i$. We use the pseudo-labels at the current iteration $t$ as the ``ground-truth''.
Then, the average pseudo-margin (APM) for the unlabeled sample $\hat{x}_i$ with pseudo-label $c$ at iteration $t$ is defined as follows:
\begin{equation}
    \text{APM}^t_{c}(\hat{x}_i) = \frac{1}{t} \sum_{e=1}^t \left( \text{PM}^e_{c}(\hat{x}_i) \right)
\vspace{-2mm}    
\end{equation}
An important consideration here is that for any previous iteration $t'$, if the pseudo-label was $c'$ (where $c$ $\neq$ $c'$) the pseudo-margin $\text{PM}^{t'}$ is calculated with respect to $c'$ and also the APM is averaged from 1 to $t'$ with respect to $c'$. In practice, we maintain a vector of pseudo-margins for all classes accumulated over the training iterations and dynamically retrieve the accumulated pseudo-margin value of the argmax class $c$ to obtain the APM$^t_c$ at iteration $t$. 

To handle old pseudo-margin deprecation across a large number of iterations, we use an exponential moving average of pseudo-margins to place higher importance on recent iterations. Hence, APM follows: 
{\footnotesize
\begin{equation}
    \text{APM}_c^t(\hat{x}_i) = \text{PM}_c^t(\hat{x}_i) \cdot \frac{\delta}{1 + t} + \text{APM}_c^{t-1}(\hat{x}_i) \cdot \left(1 - \frac{\delta}{1 + t} \right)
\end{equation}
}
Similar to \citet{sosea2023marginmatch}, we set the smoothing parameter \(\delta = 0.997\).

For both labeled and unlabeled data, samples with larger AUM and APM values, respectively, are considered easier for the model to learn, and those with smaller values are generally ambiguous, harder to learn, or mislabeled. These hard to learn or ambiguous samples are often the cause of overconfidence and errors. Hence, in the mixup, we include one easy sample and one hard sample. The easy-to-learn sample ensures that the resulting pseudo-label of the mixed sample is qualitative, while the presence of the hard-to-learn sample ensures more uncertainty for the model. To perform these difficulty-aware splits, we partition both labeled and pseudo-labeled data at each iteration into “easy-to-learn” and “hard-to-learn” subsets. This categorization is based on the batch medians of AUM for labeled samples ($\tau_L$) and APM for pseudo-labeled samples ($\tau_U$), following \cite{park2022calibration} but in a batch-wise manner, effectively distinguishing samples by their learning difficulty.

After obtaining the difficulty-aware asymmetric splits, mixup is then performed as follows: for each easy labeled sample, we randomly select one from its \textit{top-k} most dissimilar hard pseudo-labeled samples, and for each hard labeled sample, we do the same with the \textit{top-k} most dissimilar easy pseudo-labeled samples. This dissimilarity based sample selection for mixup better simulates the out-of-domain (OOD) distribution by producing mixup augmented examples that deviate from the in-domain distribution. Furthermore, randomly selecting a sample from the \textit{top-k} most dissimilar samples introduces diversity across iterations and avoids overfitting. As a result, the model becomes less likely to make overconfident predictions on ambiguous inputs, thereby improving calibration. For the mixup, we compute a convex combination of the two inputs using weights $\gamma$ and $1 - \gamma$ as follows:

\vspace{-2mm}
\begin{equation}
\vspace{-2mm}
\tilde{x_1} = \gamma x_{le} + (1 - \gamma) \hat{x}_{uh}
\end{equation}
\vspace{-2mm}
\begin{equation}
\vspace{-2mm}
\tilde{y_1} = \gamma y_{le} + (1 - \gamma) \hat{y}_{uh}
\end{equation}
\vspace{-2mm}
\begin{equation}
\vspace{-2mm}
\tilde{x_2} = \gamma x_{lh} + (1 - \gamma) \hat{x}_{ue}
\end{equation}
\vspace{-2mm}
\begin{equation}
\tilde{y_2} = \gamma y_{lh} + (1 - \gamma) \hat{y}_{ue}
\end{equation}
where \( x_{le}, y_{le} \) are easy labeled samples and their ground-truth labels, \( x_{lh}, y_{lh} \) are hard labeled samples and their ground-truth labels, \( \hat{x}_{ue}, \hat{y}_{ue} \) are easy pseudo-labeled samples and their pseudo-labels, and \( \hat{x}_{uh}, \hat{y}_{uh} \) are hard pseudo-labeled samples and their pseudo-labels.

The final model input includes labeled, unlabeled, and mixup-augmented data. The total loss $\mathcal{L}$ is computed as:
\begin{equation}
\mathcal{L} = \mathcal{L}_L + \lambda_U \cdot \mathcal{L}_U + \mathcal{L}_{\text{mixup}}
\end{equation}
where $\mathcal{L}_L$ is the supervised loss on labeled data, $\mathcal{L}_U$ is the unsupervised loss on the unlabeled data and \(\lambda_U\) is a hyperparameter that controls the weight of $\mathcal{L}_U$ in the total loss $\mathcal{L}$, and $\mathcal{L}_{\text{mixup}}$ is the loss on mixup-augmented samples. In experiments, we set \(\lambda_U\) equals 1 consistent with previous approaches \cite{sohn2020fixmatch, chen2023softmatch}.

The CalibrateMix algorithm is shown in Algorithm \ref{alg: CalibrateMix}. 

\begin{algorithm}
\begingroup
\scriptsize
\begin{algorithmic}[1]
\REQUIRE Labeled batch $D_{L}$; unlabeled batch $D_{U}$; model $\theta$; number of classes $C$; weak augmentation $\pi$; strong augmentation $\Pi$; mixup coefficient $\gamma$; number of dissimilar samples $k$; confidence threshold $\omega$; labeled-to-unlabeled loss ratio $\lambda$; medians of labeled and unlabeled batches $\tau_L$, $\tau_U$; current iteration $t$.

\STATE Compute pseudo-labels for the unlabeled batch: \newline
$\hat{y}^{t}_i = \arg\max_{c \in \{1, \dots, C\}} p_{\theta}(\pi(\hat{x}_i))$ for $\hat{x}_i \in D_{U}$ and build $\hat{D}_U = \{(\hat{x_i}, \hat{y}^{t}_{i}) \mid \hat{x_i} \in D_{U}\}$
\STATE Calculate AUM$^{t}$ for samples in $D_{L}$ and APM$^{t}$ for samples in $D_{U}$
\STATE \textbf{Split} the labeled and unlabeled batch based on samples' learning difficulty:
\STATE \quad $D_{L_{easy}} \leftarrow \{ (x_i, y_i) \in D_{L} \mid \ \text{AUM}^{t}(x_i, y_i) \geq \tau_L \}$
\STATE \quad $D_{L_{hard}} \leftarrow\{ (x_i, y_i) \in D_{L} \mid \ \text{AUM}^{t}(x_i, y_i) < \tau_L \}$
\STATE \quad $\hat{D}_{U_{easy}} \leftarrow \{ (\hat{x_i}, \hat{y}^{t}_{i}) \in \hat{D}_{U} \mid \ \text{APM}^{t}(\hat{x_i}, \hat{y}^{t}_{i}) \geq \tau_U \}$
\STATE \quad $\hat{D}_{U_{hard}} \leftarrow\{ (\hat{x_i}, \hat{y}^{t}_{i}) \in \hat{D}_{U} \mid \ \text{APM}^{t}(\hat{x_i}, \hat{y}^{t}_{i}) < \tau_U \}$

\STATE \textbf{Mix} easy labeled with hard unlabeled and hard labeled with easy unlabeled following Eqs (6), (7), (8), (9):
\STATE \quad \hspace{0.4em} $\tilde{D}_{{Mix1}} \gets \text{Mixup}(D_{L_{easy}}, \textit{top-k dissimilar samples from } \hat{D}_{U_{hard}})$
\STATE \quad \hspace{0.38em} $\tilde{D}_{Mix2} \gets \text{Mixup}(D_{L_{hard}}, \textit{top-k dissimilar samples from } \hat{D}_{U_{easy}})$
\STATE \quad $D_{M} \leftarrow \tilde{D}_{Mix1} + \tilde{D}_{Mix2}$

\STATE \textbf{Optimize} total loss: $\mathcal{L} \leftarrow \mathcal{L}_{L} + \lambda \cdot \mathcal{L}_{U} + \mathcal{L}_{Mixup}$, where
\STATE $\mathcal{L}_{L} = \frac{1}{|D_L|} \sum_{k=1}^{|D_L|} H(y_k,p_{\theta}(\pi(x_k)))$
\STATE $\mathcal{L}_{U} = \frac{1}{|D_U|} \sum_{k=1}^{|D_U|} \mathds{1}(\max(p_\theta(\pi(\hat{x}_k))) \geq \omega) \cdot H(\hat{y}^{t}_k,\ p_\theta(\Pi(\hat{x}_k)))$
\STATE $\mathcal{L}_{Mixup} = \frac{1}{|D_M|} \sum_{k=1}^{|D_M|} H(\tilde{y}_{k}, p_\theta(\tilde{x}_k))$

\end{algorithmic}
\caption{CalibrateMix}
\label{alg: CalibrateMix}
\endgroup
\end{algorithm}

\section{Experimental Analysis}

\label{sec:exp}
We test the performance of our approach across a variety of image benchmarks by following standard SSL settings \cite{chen2023softmatch,NEURIPS2022_190dd6a5}. Specifically, we conduct experiments varying the amounts of labeled samples on CIFAR-10, CIFAR-100 \cite{krizhevsky2009learning}, SVHN \cite{netzer2011reading}, STL-10 \cite{pmlr-v15-coates11a}, ImageNet \cite{deng2009imagenet}, and WebVision \cite{li2017webvision}. For the CIFAR-10, CIFAR-100, SVHN, and STL-10 datasets, we follow \citet{sohn2020fixmatch} and randomly select a small number of labeled samples ranging from $4$ labels per class up to $400$ labels per class and treat the remaining samples as unlabeled data, except for STL-10, which has its own set of unlabeled samples. For the ImageNet and WebVision datasets, we use 10\% of the available labeled samples as labeled data and the remaining 90\% as unlabeled data. In addition to the SSL setups, we also show results on the Fully Supervised setting.

\begin{table*}[ht]

\centering
\small
\resizebox{\textwidth}{!}{
\begin{tabular}{p{1.8cm}|l|ccc|ccc|cc|cc}
\toprule
\textbf{Metric} & \multicolumn{1}{c|}{\textbf{Method}} &
\multicolumn{3}{c|}{\textbf{CIFAR-10}} &
\multicolumn{3}{c|}{\textbf{CIFAR-100}} &
\multicolumn{2}{c|}{\textbf{SVHN}} &
\multicolumn{2}{c}{\textbf{STL-10}} \\
\cmidrule(lr){3-5} \cmidrule(lr){6-8} \cmidrule(lr){9-10} \cmidrule(lr){11-12}
& & \textbf{4} & \textbf{25} & \textbf{400} & \textbf{4} & \textbf{25} & \textbf{100} & \textbf{4} & \textbf{100} & \textbf{4} & \textbf{100} \\
\midrule

\multirow{2}{*}{\textbf{ECE}} 
& \textbf{FixMatch} 
& $5.42_{0.36}$ & $2.85_{0.23}$ & $2.68_{0.09}$ 
& $35.88_{1.1}$ & $21.19_{0.62}$ & $15.73_{0.05}$ 
& $\mathbf{1.02}_{0.02}$ & $1.1_{0.04}$ 
& $32.66_{1.29}$ & $4.22_{0.37}$ \\
& \textbf{FixMatch + CalibrateMix (Ours)} 
& $\mathbf{3.24}_{0.12}$ & $\mathbf{1.97}_{0.13}$ & $\mathbf{1.25}_{0.27}$ 
& $\mathbf{35.43}_{0.05}$ & $\mathbf{12.68}_{0.42}$ & $\mathbf{8.74}_{0.6}$ 
& $1.06_{0.07}$ & $\mathbf{0.82}_{0.13}$ 
& $\mathbf{29.93}_{0.88}$ & $\mathbf{3.9}_{0.25}$ \\
\midrule

\multirow{2}{*}{\textbf{ECE}} 
& \textbf{FlexMatch} 
& $4.76_{0.95}$ & $\mathbf{2.92}_{0.53}$ & $2.98_{0.55}$ 
& $31.51_{9.55}$ & $29.54_{7.24}$ & $18.55_{4.15}$ 
& $10.15_{2.88}$ & $7.81_{1.55}$ 
& $35.62_{8.57}$ & $9.45_{2.12}$ \\
& \textbf{FlexMatch + CalibrateMix (Ours)} 
& $\mathbf{4.41}_{0.22}$ & $2.94_{0.15}$ & $\mathbf{2.95}_{0.24}$ 
& $\mathbf{27.65}_{0.33}$ & $\mathbf{28.23}_{0.27}$ & $\mathbf{18.02}_{0.54}$ 
& $\mathbf{10.01}_{0.08}$ & $\mathbf{7.54}_{0.11}$ 
& $\mathbf{32.45}_{0.23}$ & $\mathbf{9.26}_{0.41}$ \\
\midrule

\multirow{2}{*}{\textbf{ECE}} 
& \textbf{SoftMatch} 
& $3.02_{0.21}$ & $2.95_{0.53}$ & $2.21_{0.03}$ 
& $26.4_{1.2}$ & $19.58_{0.1}$ & $15.34_{0.32}$ 
& $1.44_{0.03}$ & $1.17_{0.07}$ 
& $13.24_{0.06}$ & $4.31_{0.4}$ \\
& \textbf{SoftMatch + CalibrateMix (Ours)} 
& $\mathbf{1.69}_{0.15}$ & $\mathbf{2.09}_{0.18}$ & $\mathbf{1.40}_{0.26}$ 
& $\mathbf{23.1}_{0.3}$ & $\mathbf{13.9}_{0.1}$ & $\mathbf{11.9}_{0.33}$ 
& $\mathbf{0.70}_{0.05}$ & $\mathbf{0.80}_{0.03}$ 
& $\mathbf{12.4}_{0.08}$ & $\mathbf{3.72}_{0.6}$ \\
\midrule

\multirow{2}{*}{\textbf{ECE}} 
& \textbf{FullySupervised} 
& \multicolumn{3}{c|}{$2.22_{0.13}$} 
& \multicolumn{3}{c|}{$6.77_{0.67}$} 
& \multicolumn{2}{c|}{$0.4_{0.04}$} 
& \multicolumn{2}{c}{$24.14_{0.38}$} \\
& \textbf{FullySupervised + CalibrateMix (Ours)} 
& \multicolumn{3}{c|}{$\mathbf{1.84}_{0.22}$} 
& \multicolumn{3}{c|}{$\mathbf{4.74}_{0.13}$} 
& \multicolumn{2}{c|}{$\mathbf{0.29}_{0.11}$} 
& \multicolumn{2}{c}{$\mathbf{13.77}_{0.23}$} \\
\midrule

\multirow{2}{*}{\textbf{Error Rate}} 
& \textbf{FixMatch} 
& $7.29_{0.05}$ & $\mathbf{4.91}_{0.02}$ & $4.3_{0.02}$ 
& $44.45_{0.15}$ & $29.88_{0.17}$ & $22.88_{0.03}$ 
& $3.65_{0.23}$ & $\mathbf{2.04}_{0.1}$ 
& $\mathbf{36.34}_{0.4}$ & $6.2_{0.07}$ \\
& \textbf{FixMatch + CalibrateMix (Ours)} 
& $\mathbf{7.13}_{0.02}$ & $4.97_{0.01}$ & $\mathbf{4.26}_{0.08}$ 
& $\mathbf{44.05}_{0.11}$ & $\mathbf{28.34}_{0.41}$ & $\mathbf{22.3}_{0.3}$ 
& $\mathbf{3.63}_{0.11}$ & $2.17_{0.03}$ 
& $37.3_{0.12}$ & $\mathbf{5.78}_{0.06}$ \\
\midrule

\multirow{2}{*}{\textbf{Error Rate}} 
& \textbf{FlexMatch} 
& $\mathbf{5.03}_{0.08}$ & $\mathbf{4.98}_{0.07}$ & $4.28_{0.02}$ 
& $40.15_{1.87}$ & $27.73_{0.35}$ & $21.93_{0.33}$ 
& $7.88_{1.23}$ & $6.78_{1.15}$ 
& $29.78_{4.01}$ & $6.29_{0.54}$ \\
& \textbf{FlexMatch + CalibrateMix (Ours)} 
& $5.04_{0.05}$ & $4.99_{0.02}$ & $\mathbf{4.23}_{0.05}$ 
& $\mathbf{40.02}_{0.16}$ & $\mathbf{26.33}_{0.31}$ & $\mathbf{21.55}_{0.16}$ 
& $\mathbf{7.56}_{0.14}$ & $\mathbf{6.71}_{0.08}$ 
& $\mathbf{29.03}_{0.43}$ & $\mathbf{6.25}_{0.09}$ \\
\midrule 

\multirow{2}{*}{\textbf{Error Rate}} 
& \textbf{SoftMatch} 
& $5.09_{0.07}$ & $\mathbf{4.90}_{0.02}$ & $4.16_{0.07}$ 
& $37.11_{0.3}$ & $\mathbf{26.76}_{0.1}$ & $\mathbf{22.11}_{0.3}$ 
& $2.59_{0.8}$ & $2.09_{0.02}$ 
& $20.90_{2.47}$ & $6.10_{0.06}$ \\
& \textbf{SoftMatch + CalibrateMix (Ours)} 
& $\mathbf{5.03}_{0.06}$ & $4.93_{0.03}$ & $\mathbf{4.05}_{0.04}$ 
& $\mathbf{36.68}_{0.18}$ & $26.86_{0.13}$ & $22.21_{0.03}$ 
& $\mathbf{2.58}_{0.01}$ & $\mathbf{2.08}_{0.2}$ 
& $\mathbf{15.18}_{0.63}$ & $\mathbf{6.07}_{0.04}$ \\
\midrule

\multirow{2}{*}{\textbf{Error Rate}} 
& \textbf{FullySupervised} 
& \multicolumn{3}{c|}{$4.63_{0.03}$} 
& \multicolumn{3}{c|}{$19.42_{0.17}$} 
& \multicolumn{2}{c|}{$2.19_{0.01}$} 
& \multicolumn{2}{c}{$34.41_{0.17}$} \\
& \textbf{FullySupervised + CalibrateMix (Ours)} 
& \multicolumn{3}{c|}{$\mathbf{4.51}_{0.08}$} 
& \multicolumn{3}{c|}{$\mathbf{18.97}_{0.26}$} 
& \multicolumn{2}{c|}{$\mathbf{2.17}_{0.04}$} 
& \multicolumn{2}{c}{$\mathbf{27.81}_{0.21}$} \\
\bottomrule
\end{tabular}
}
\caption{Expected Calibration Errors (ECE, \%) and Top-1 error rates (\%) on CIFAR-10, CIFAR-100, SVHN, and STL-10 datasets by FixMatch, FlexMatch, SoftMatch and CalibrateMix (the lower the better). Values are reported in the format $X_Y$ where $X$ is the mean and $Y$ is the standard deviation across 3 runs. Better scores in comparison are shown in \textbf{bold}.}
\label{tab:table1}

\end{table*}

\begin{table*}[ht]

\centering

\resizebox{\textwidth}{!}{
\begin{tabular}{p{1.8cm}|l|ccc|ccc|cc|cc}
\toprule
\textbf{Metric} & \multicolumn{1}{c|}{\textbf{Method}} &
\multicolumn{3}{c|}{\textbf{CIFAR-10}} &
\multicolumn{3}{c|}{\textbf{CIFAR-100}} &
\multicolumn{2}{c|}{\textbf{SVHN}} &
\multicolumn{2}{c}{\textbf{STL-10}} \\
\cmidrule(lr){3-5} \cmidrule(lr){6-8} \cmidrule(lr){9-10} \cmidrule(lr){11-12}
& & \textbf{4} & \textbf{25} & \textbf{400} & \textbf{4} & \textbf{25} & \textbf{100} & \textbf{4} & \textbf{100} & \textbf{4} & \textbf{100} \\
\midrule

\multirow{1}{*}{\textbf{ECE}} 
& \textbf{FixMatch + RandomMixup} & 5.26$_{2.61}$ & 3.19$_{1.42}$ & 1.72$_{0.11}$ & 38.05$_{2.51}$ & 14.78$_{0.21}$ & \textbf{8.72}$_{1.91}$ & 2.87$_{0.25}$ & 0.87$_{0.06}$ & 35.58$_{3.61}$ & 4.16$_{0.30}$ \\

& \textbf{FixMatch + CalibrateMix (Ours)} & \textbf{3.24$_{0.12}$} & \textbf{1.97$_{0.13}$} & \textbf{1.25$_{0.27}$} & \textbf{35.43$_{0.05}$} & \textbf{12.68$_{0.42}$} & 8.74$_{0.6}$ & \textbf{1.06$_{0.07}$} & \textbf{0.82$_{0.13}$} & \textbf{29.93$_{0.88}$} & \textbf{3.9$_{0.25}$}  \\

& \textbf{FixMatch + LS} & 4.07$_{0.48}$ & 2.52$_{0.22}$ & 2.69$_{0.11}$ & 34.11$_{2.01}$ & 17.37$_{1.11}$ & 7.56$_{0.95}$ & 1.23$_{0.05}$ & 1.19$_{0.06}$ & 32.96$_{1.73}$ & 4.02$_{0.32}$ \\

& \textbf{FixMatch + CalibrateMix (Ours) + LS} & \textbf{4.07$_{0.41}$} &\textbf{ 2.06$_{0.23}$} & \textbf{1.22$_{0.45}$} & \textbf{31.7$_{1.33}$} & \textbf{12.89$_{1.1}$} & \textbf{7.11$_{0.8}$} & \textbf{1.22$_{0.1}$} & \textbf{0.99$_{0.13}$} & \textbf{29.56$_{2.34}$} & \textbf{3.9$_{0.28}$} \\

\midrule

\multirow{1}{*}{\textbf{ECE}} 
& \textbf{FlexMatch + RandomMixup} & 4.52$_{0.18}$ & 3.57$_{0.31}$ & 3.24$_{0.02}$ & 27.80$_{0.08}$ & \textbf{26.72}$_{1.05}$ & 21.79$_{0.19}$ & 10.45$_{0.34}$ & 7.80$_{0.76}$ & \textbf{31.80}$_{1.38}$ & 9.35$_{0.23}$ \\

& \textbf{FlexMatch + CalibrateMix (Ours)} & \textbf{4.41$_{0.22}$} & \textbf{2.94$_{0.15}$} & \textbf{2.95$_{0.24}$} & \textbf{27.65$_{0.33}$} & 28.23$_{0.27}$ & \textbf{18.02$_{0.54}$} & \textbf{10.01$_{0.08}$} & \textbf{7.54$_{0.11}$} & 32.45$_{0.23}$ & \textbf{9.26$_{0.41}$}  \\

& \textbf{FlexMatch + LS} & 4.01$_{0.32}$ & \textbf{2.23$_{0.21}$} & 2.96$_{0.31}$ & 29.41$_{5.71}$ & 29.03$_{3.17}$ & 18.17$_{3.61}$ & \textbf{9.65$_{1.16}$} & 7.23$_{1.34}$ & 31.87$_{4.5}$ & 7.86$_{1.54}$ \\

& \textbf{FlexMatch + CalibrateMix (Ours) + LS} & \textbf{3.99$_{0.32}$} & 2.24$_{0.19}$ & \textbf{2.91$_{0.14}$} & \textbf{26.84$_{1.34}$} & \textbf{27.53$_{01.05}$} & \textbf{17.03$_{0.61}$} & 9.67$_{0.15}$ & \textbf{7.01$_{0.16}$} & \textbf{30.44$_{1.73}$} & \textbf{7.1$_{0.34}$} \\
\midrule
\multirow{1}{*}{\textbf{ECE}} 
& \textbf{SoftMatch + RandomMixup} & 2.75$_{0.64}$ & 2.17$_{0.10}$ & 1.84$_{0.04}$ & 26.65$_{1.12}$ & 14.12$_{0.32}$ & \textbf{11.62}$_{0.58}$ & 2.44$_{0.28}$ & 0.84$_{0.05}$ & 21.71$_{3.47}$ & 4.07$_{0.05}$ \\

& \textbf{SoftMatch + CalibrateMix (Ours)} & \textbf{1.69$_{0.15}$} & \textbf{2.09$_{0.18}$} & \textbf{1.4$_{0.26}$} & \textbf{23.1$_{0.3}$} & \textbf{13.9$_{0.1}$} & 11.9$_{0.33}$ & \textbf{0.7$_{0.05}$} & \textbf{0.8$_{0.03}$} & \textbf{12.4$_{0.08}$} & \textbf{3.72$_{0.6}$}  \\

& \textbf{SoftMatch + LS} & 6.62$_{0.32}$ & 6.83$_{0.3}$ & 7.02$_{0.08}$ & 20.89$_{0.40}$ & 7.88$_{0.37}$ & 6.11$_{0.35}$ & 8.31$_{0.34}$ & 7.80$_{0.23}$ & 11.26$_{3.44}$ & 7.46$_{0.5}$ \\

& \textbf{SoftMatch + CalibrateMix (Ours) + LS} & \textbf{4.76$_{0.39}$} & \textbf{5.70$_{0.37}$} & \textbf{6.49$_{0.11}$} & \textbf{16.27$_{0.44}$} & \textbf{6.25$_{0.59}$} & \textbf{4.49$_{0.19}$} & \textbf{4.64$_{1.49}$} & \textbf{7.32$_{0.06}$} & \textbf{8.05$_{2.18}$} & \textbf{7.01$_{0.25}$} \\
\midrule

\multirow{1}{*}{\textbf{Error Rate}} 
& \textbf{FixMatch + RandomMixup} & 8.98$_{1.85}$ & 6.89$_{1.33}$ & 4.35$_{0.04}$ & 50.53$_{2.19}$ & \textbf{27.17$_{0.33}$} & \textbf{20.76$_{0.23}$} & \textbf{2.48$_{0.04}$} & 2.40$_{0.13}$ & 40.69$_{5.30}$ & 6.44$_{0.11}$ \\

& \textbf{FixMatch + CalibrateMix (Ours)} & \textbf{7.13$_{0.02}$} & \textbf{4.97$_{0.01}$} & \textbf{4.26$_{0.08}$} & \textbf{44.05$_{0.11}$} & 28.34$_{0.41}$ & 22.3$_{0.3}$ & 3.63$_{0.11}$ & \textbf{2.17$_{0.03}$} & \textbf{37.3$_{0.12}$} & \textbf{5.78$_{0.06}$}  \\

& \textbf{FixMatch + LS} & 7.41$_{0.61}$ & 5.06$_{0.98}$ & 4.32$_{0.17}$ & 44.57$_{4.16}$ & 29.82$_{2.64}$ & 23.62$_{1.44}$ & 3.71$_{0.46}$ & 2.11$_{0.24}$ & \textbf{36.43$_{2.32}$} & 6.5$_{0.28}$ \\

& \textbf{FixMatch + CalibrateMix (Ours) + LS} & \textbf{7.13$_{0.61}$} & \textbf{4.99$_{0.23}$} & \textbf{4.27$_{0.25}$} &\textbf{ 44.11$_{1.42}$} & \textbf{28.64$_{1.24}$} & \textbf{22.52$_{1.05}$} & \textbf{3.64$_{0.31}$} & \textbf{2.11$_{0.25}$} & 37.54$_{1.76}$ & \textbf{5.72$_{0.42}$} \\

\midrule
\multirow{1}{*}{\textbf{Error Rate}} 
& \textbf{FlexMatch + RandomMixup} & 5.84$_{0.34}$ & 5.45$_{0.08}$ & 4.69$_{0.05}$ & 40.62$_{0.88}$ & 26.53$_{0.43}$ & 22.12$_{0.13}$ & 7.78$_{0.45}$ & 6.96$_{0.36}$ & 31.60$_{6.21}$ & 6.66$_{0.13}$ \\

& \textbf{FlexMatch + CalibrateMix (Ours)} & \textbf{5.04$_{0.05}$} & \textbf{4.99$_{0.02}$} & \textbf{4.23$_{0.05}$} & \textbf{40.02$_{0.16}$} & \textbf{26.33$_{0.31}$} & \textbf{21.55$_{0.16}$} & \textbf{7.56$_{0.14}$} & \textbf{6.71$_{0.08}$} & \textbf{29.03$_{0.43}$} & \textbf{6.25$_{0.09}$}  \\

& \textbf{FlexMatch + LS} & \textbf{4.81$_{0.51}$} & 4.72$_{0.47}$ & \textbf{4.01$_{0.34}$} & 40.11$_{1.64}$ & 27.56$_{1.15}$ & 21.53$_{1.04}$ & 7.51$_{0.31}$ & 6.91$_{0.37}$ & 29.44$_{1.27}$ & \textbf{5.88$_{0.49}$} \\

& \textbf{FlexMatch + CalibrateMix (Ours) + LS} & 4.83$_{0.41}$ & \textbf{4.71$_{0.37}$} & 4.05$_{0.22}$ & \textbf{40.01$_{1.87}$} & \textbf{26.02$_{1.46}$} & \textbf{21.35$_{1.21}$} & \textbf{7.45$_{0.48}$} & \textbf{6.63$_{0.32}$} & \textbf{28.66$_{0.28}$} & 6.04$_{0.16}$ \\

\midrule 

\multirow{1}{*}{\textbf{Error Rate}} 
& \textbf{SoftMatch + RandomMixup} & 7.44$_{0.96}$ & 6.02$_{0.36}$ & 4.42$_{0.08}$ & 40.20$_{0.6}$ & \textbf{25.42$_{0.15}$} & \textbf{21.58$_{0.86}$}& \textbf{2.51$_{0.07}$} & 2.64$_{0.18}$ & 28.74$_{6.26}$ & \textbf{5.78}$_{0.14}$ \\

& \textbf{SoftMatch + CalibrateMix (Ours)} & \textbf{5.03$_{0.06}$} & \textbf{4.93$_{0.03}$} & \textbf{4.05$_{0.04}$} & \textbf{36.68$_{0.18}$} & 26.86$_{0.13}$ & 22.21$_{0.03}$ & 2.58$_{0.01}$ & \textbf{2.08$_{0.2}$} & \textbf{15.18$_{0.63}$} & 6.07$_{0.04}$ \\

& \textbf{SoftMatch + LS} & 5.56$_{0.46}$ & 5.19$_{0.12}$ & 4.46$_{0.09}$ & 39.10$_{0.46}$ & \textbf{26.50$_{0.25}$} & 22.13$_{0.11}$ & 3.14$_{0.27}$ & 3.07$_{0.20}$ & 24.57$_{3.93}$ & \textbf{5.63$_{0.08}$} \\

& \textbf{SoftMatch + CalibrateMix (Ours) + LS} & \textbf{5.49$_{0.34}$} & \textbf{5.09$_{0.14}$} & \textbf{4.43$_{0.11}$} & \textbf{37.61$_{1}$} & 27.41$_{1.18}$ & \textbf{21.81$_{0.28}$} & \textbf{2.9$_{0.18}$} & \textbf{2.84$_{0.07}$} & \textbf{19.24$_{5.76}$} & 5.97$_{0.12}$ \\
\bottomrule

\end{tabular}
}

\caption{Expected Calibration Errors (ECE, \%) and Top-1 error rates (\%) on CIFAR-10, CIFAR-100, SVHN, and STL-10 datasets by Random Mixup, Label Smoothing (LS), and CalibrateMix (the lower the better). Values are reported in the format $X_Y $where $X$ is the mean and $Y$ is the Standard Deviation across 3 runs. Better scores in comparison are shown in \textbf{bold}.}

\label{tab:table2}
\end{table*}

We run all our experiments three times and report the average ECE and error rates, as well as their standard deviations.  Similar to \citet{sohn2020fixmatch}, we utilize Wide Residual Networks \cite{zagoruyko2016wide} for small-scale datasets: WRN-28-2 for CIFAR-10 and SVHN, WRN-37-2 for STL-10 and WRN-28-8 for CIFAR-100; and use ResNet-50 \cite{he2016deep} for the large-scale ImageNet and WebVision. Additionally, we utilize the SGD optimizer with a momentum of 0.9 and an initial learning rate of 0.03. We use a cosine annealing schedule to dynamically adjust the learning rate over a total of $2^{20}$ training steps. For each dataset, the batch size for labeled data and mixup data is set to 64, while the batch size for unlabeled data is configured to be seven times larger than that of the labeled data. We also use weak (flip and shift) and strong (RandAugment \cite{cubuk2020randaugment}) augmentations for the unlabeled data. We set the mixup coefficient $\gamma$ to be $0.4$, similar to prior work \cite{thulasidasan2019mixup}. For CalibrateMix, we include warmup during training, with the warmup phase consisting of approximately 100 epochs.

\subsection{Performance on CIFAR-10, CIFAR-100, STL-10, SVHN}

We compare CalibrateMix to relevant SSL approaches: FixMatch \cite{sohn2020fixmatch}, FlexMatch \cite{zhang2021flexmatch}, SoftMatch \cite{chen2023softmatch}, and report ECE and top-1 error rates in Table \ref{tab:table1}. We further compare CalibrateMix against two calibration techniques: Random Mixup and Label Smoothing (LS) in Table \ref{tab:table2}. For Random Mixup, we randomly pair two samples from the combined pool of labeled and unlabeled data. Additionally, we apply LS following \citet{muller2019does}, with a factor of $\gamma = 0.1$ to the labeled, unlabeled, and mixup samples for all methods: FixMatch, FlexMatch, SoftMatch, and CalibrateMix (including combining CalibrateMix with LS). 
The results for Label Smoothing are denoted as `LS' and the results with Random Mixup are denoted as `RandomMixup' in Table \ref{tab:table2}. 

Across all datasets, CalibrateMix consistently improves model calibration and, in many cases, enhances classification accuracy as can be seen from Table \ref{tab:table1}. For example, when integrated with FixMatch, CalibrateMix yields better-calibrated models. On CIFAR-10 with only $4$ labels per class, CalibrateMix achieves a reduction in ECE by $2.18\%$. On the more challenging CIFAR-100 dataset with $25$ labels per class, the improvements are more significant, reducing ECE by $8.51\%$ and error rate by $1.54\%$. On the more realistic STL-10 dataset, CalibrateMix reduces ECE by $2.73\%$, further demonstrating its robustness. CalibrateMix also significantly enhances FlexMatch, particularly under low-label settings. 
Furthermore, integrating CalibrateMix with SoftMatch provides significant gains. On CIFAR-100 with $25$ labels per class, ECE is improved by $5.68\%$.
On STL-10 with $4$ labels per class, CalibrateMix achieves a $0.84\%$ lower ECE and a $5.72\%$ reduction in error rate compared to SoftMatch. 
We can also observe that CalibrateMix yields well calibrated models in the Fully Supervised setting.

CalibrateMix consistently demonstrates improvements in model calibration and often in error rates when compared to prior calibration strategies such as random mixup and label smoothing, as shown in Table \ref{tab:table2}. When applied to FixMatch, CalibrateMix outperforms random mixup across several settings. Notably, on CIFAR-100 and STL-10 with $4$ labels per class, it achieves ECE reductions of $2.62\%$ and $5.65\%$, along with error rate reductions of $6.48\%$ and $3.39\%$, respectively. Furthermore, on CIFAR-100 with $25$ labels per class, combining CalibrateMix with label smoothing results in a $4.48\%$ reduction in ECE over using label smoothing alone with FixMatch. On FlexMatch, similar benefits are observed. 
When integrated with SoftMatch, CalibrateMix continues to enhance calibration by outperforming random mixup. For example, with $4$ labels per class on CIFAR-100 and STL-10, CalibrateMix reduces ECE by $3.55\%$ and $9.31\%$, and decreases error rates by $3.52\%$ and $13.56\%$, respectively, compared to random mixup. Compared to label smoothing, ECE is reduced by $4.62\%$ on CIFAR-100 and $3.21\%$ on STL-10 on the $4$ labels per class setting. These consistent reductions across different models and datasets highlight CalibrateMix as a strong solution to existing calibration techniques, making it an effective and versatile regularization method for SSL.

\subsection{Performance on ImageNet and WebVision}
We report the performance of CalibrateMix on two large-scale datasets: ImageNet \cite{deng2009imagenet} and WebVision \cite{li2017webvision}. We randomly sample $10\%$ examples from the training set to be used as labeled samples and use the rest of the examples as unlabeled data. We show the results of FixMatch, FlexMatch, SoftMatch and FullySupervised in terms of ECE and Error rate in Table \ref{tab:imagenet_webvision}. We note that CalibrateMix considerably boosts the calibration of our models in all setups and yields small error rate improvements in most settings. Specifically, CalibrateMix outperforms SoftMatch by $1.79\%$ ECE on WebVision and pushes the performance over FixMatch by $0.99\%$ ECE on ImageNet. Additionally, CalibrateMix reduces ECE by $1.98\%$ over FlexMatch on the WebVision dataset. When applied to the FullySupervised setting, it also reduces ECE by $1.50\%$ compared to the base supervised model on WebVision. These results highlight the effectiveness of our approach in large-scale, real-world scenarios, leading to better-calibrated models.

\begin{table}[h]

\centering
\small 
\resizebox{0.9\columnwidth}{!}{
\begin{tabular}{l|cc|cc}
\toprule
\textbf{Dataset} & \multicolumn{2}{c|}{\textbf{ImageNet}} & \multicolumn{2}{c}{\textbf{WebVision}} \\
& \textbf{ECE} & \textbf{Error} & \textbf{ECE} & \textbf{Error} \\
\midrule
\textbf{Fixmatch} & 9.44 & \textbf{43.54} & 12.5 & 44.51 \\
\textbf{Fixmatch + Ours} & \textbf{8.45} & 43.61 & \textbf{10.4} & \textbf{44.47} \\
\midrule
\textbf{Flexmatch} & 10.45 & \textbf{42.9} & 13.32 & 43.7 \\
\textbf{Flexmatch + Ours} & \textbf{9.76} & 43.01 & \textbf{11.34} & \textbf{43.66} \\
\midrule
\textbf{SoftMatch} & 9.88 & 40.05 & 10.8 & 42.01 \\
\textbf{SoftMatch + Ours} & \textbf{7.56} & \textbf{39.88} & \textbf{9.01} & \textbf{41.77} \\
\midrule
\textbf{FullySupervised} & 6.55 & \textbf{21.06} & 7.57 & 27.54 \\
\textbf{FullySupervised + Ours} & \textbf{5.76} & 21.87 & \textbf{6.07} & \textbf{27.22} \\
\bottomrule
\end{tabular}
}
\caption{ECE and Top-1 error rates on ImageNet and WebVision (the lower the better). Better results in comparison are shown in {\bf bold}.}
\label{tab:imagenet_webvision}
\end{table}

\section{Ablation Study}
\label{sec:ablation}
We conduct experiments to capture: (1) the effect of the targeted mixup on different setups such as how to mix labeled and unlabeled samples as well as with and without warmup, and (2) mixup with and without considering the dissimilar samples (without considering cosine similarity). 

\subsection{Effect of Different Mixup Strategies with and without Warmup}

We perform this experiment on CIFAR-100 and STL-10 ($4$ labels per class setup) using SoftMatch and the results are shown in Table \ref{tab:ablation_1}. We denote ``easy to learn" labeled samples as \textit{LE}, ``hard to learn" labeled samples as \textit{LH}, ``easy to learn" unlabeled samples as \textit{UE}, and ``hard to learn" unlabeled samples as \textit{UH}. In our first ablation, we compare CalibrateMix that mixes in between Unlabeled and Labeled ``easy to learn" and ``hard to learn" samples \textit{(LE+UH, LH+UE)} with mixup that mixes not only in between Labeled and Unlabeled sets but also within Labeled and within Unlabeled sets individually \textit{(LE+LH, LE+UH, LH+UE, UE+UH)}. We compare this setup with and without warmup and denote it as mixup all +/-warmup. 

As we can see from the table, CalibrateMix reduces both ECE and error rate compared with mixup all +warmup. Comparing mixup all with and without warmup, we can see that without warmup results in high error rates for both CIFAR-100 and STL-10 (although lower ECE on CIFAR-100).


\begin{table}[t]

\centering
\resizebox{\columnwidth}{!}{
\begin{tabular}{l|cc|cc}
\toprule
\textbf{Method} & \multicolumn{2}{c|}{\textbf{CIFAR-100}} & \multicolumn{2}{c}{\textbf{STL-10}} \\
& \textbf{ECE} & \textbf{Err} & \textbf{ECE} & \textbf{Err} \\
\midrule
\textbf{SoftMatch} & 26.4\textsubscript{1.2} & 37.11\textsubscript{0.3} & 13.24\textsubscript{0.06} & 20.90\textsubscript{2.47} \\

+ \textbf{mixup all -warmup} & 16.5\textsubscript{3.35} & 86.95\textsubscript{1.05} & 18.77\textsubscript{2.21} & 23.41\textsubscript{2.42} \\
+ \textbf{mixup all +warmup} & 30.63\textsubscript{0.64} & 41.5\textsubscript{0.76} & 16.66\textsubscript{3.35} & 19.70\textsubscript{5.07} \\

+ \textbf{CM k=0 (no cosine)} & 23.25\textsubscript{0.32} & 37.13\textsubscript{0.35} & 16.81\textsubscript{4.01} & 15.69\textsubscript{2.78} \\

+ \textbf{CM k=10} & 24.42\textsubscript{1.04} & 37.8\textsubscript{1.3} & 12.71\textsubscript{1.02} & 14.92\textsubscript{1.03} \\
+ \textbf{CM k=15} & 23.14\textsubscript{0.04} & 36.92\textsubscript{0.02} & 14.66\textsubscript{0.71} & 15.38\textsubscript{0.85} \\

+ \textbf{CM k=5 (Ours)} & 23.1\textsubscript{0.3} & 36.68\textsubscript{0.18} & 12.4\textsubscript{0.08} & 15.18\textsubscript{0.63} \\
\bottomrule
\end{tabular}
}
\caption{Ablation on CIFAR-100 and STL-10 (4 labels/class). CalibrateMix has warmup and cosine similarity. CM stands for CalibrateMix.}
\label{tab:ablation_1}
\end{table}

\subsection{Effect of Selecting Dissimilar Samples}

\label{sec:error_analysis}
\begin{figure}[t]
    \centering
    \includegraphics[width=\columnwidth]{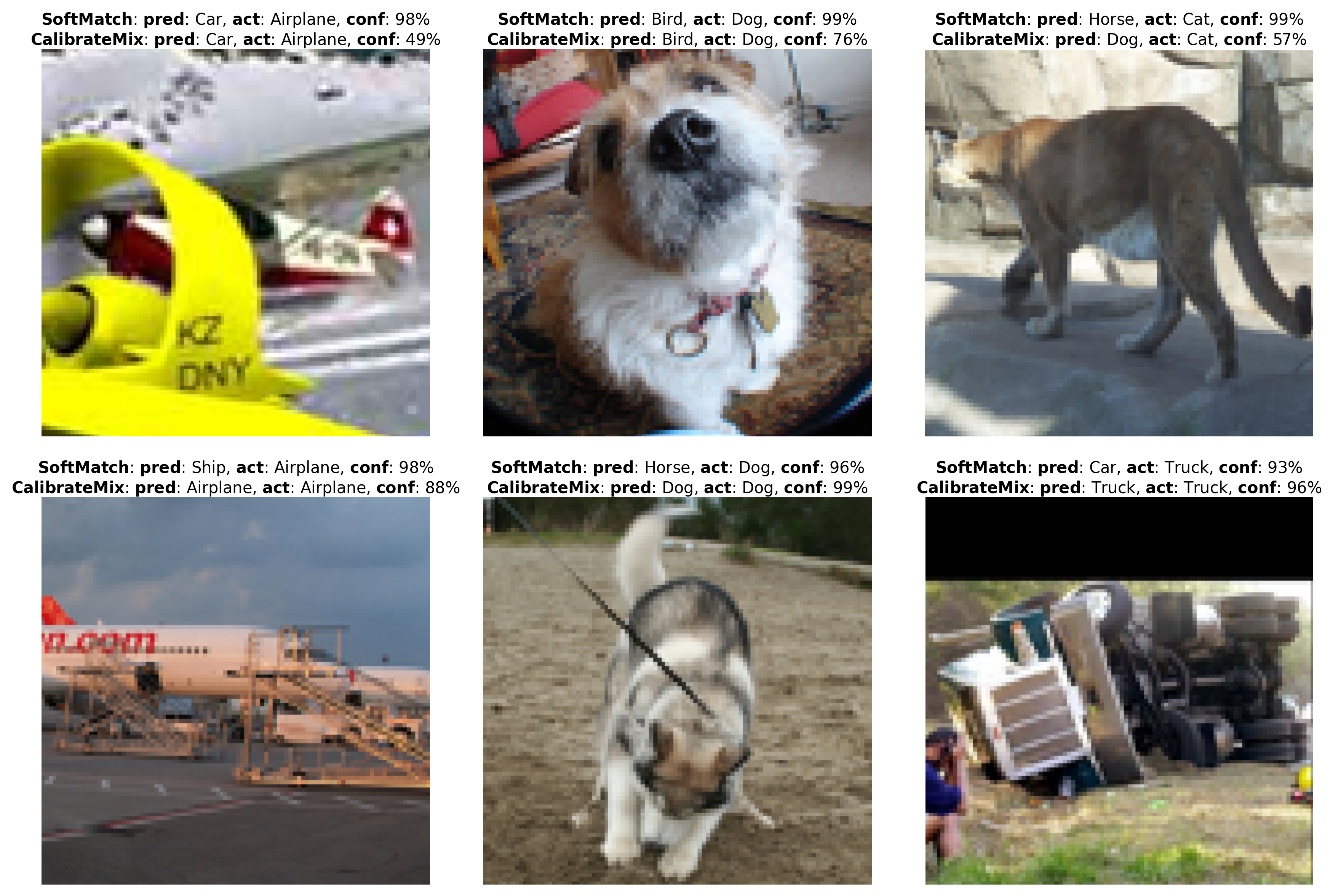}
    \caption{Confidence comparison of images of the STL-10 dataset by SoftMatch and CalibrateMix}
    \label{fig:fig_error_analysis}
\end{figure}

We further conduct experiments to show our justification behind the selection of the \textit{top-k} most dissimilar samples for mixup. To test our approach, we perform experiments on SoftMatch + CalibrateMix with different values of \textit{k}: $5\%$, $10\%$, $15\%$ dissimilar samples. We conduct experiments on CIFAR-100 and STL-10 and report our results in Table \ref{tab:ablation_1}. Our findings show that the best results are obtained with cosine similarity for \textit{k}=$5\%$ and that not using the cosine similarity increases ECE and error rate, especially on STL-10.

\section{Error Analysis}

Figure \ref{fig:fig_error_analysis} shows the confidence, predicted, and actual labels for six STL-10 test images under the 4 labels per class setting. In the top row, both CalibrateMix and SoftMatch make incorrect predictions; however, SoftMatch makes predictions with very high (over 98\%) confidence, while CalibrateMix assigns lower confidence, expressing appropriate uncertainty and thus better calibration. In the bottom row, SoftMatch once again makes high-confidence misclassification, whereas CalibrateMix correctly predicts with high confidence, further demonstrating its superior calibration and reliability. These examples illustrate that CalibrateMix not only reduces overconfidence in incorrect predictions but also preserves high certainty/high confidence in the correct ones.

\section{Conclusion}
\label{sec:conclusion}
We proposed CalibrateMix, a targeted mixup strategy that improves the calibration of SSL models by utilizing training dynamics and dissimilarity-aware pairing of easy and hard samples. Specifically, CalibrateMix leverages Area Under the Margin (AUM) and Average Pseudo Margin (APM) to identify sample difficulty, and then performs mixup between labeled and pseudo-labeled samples based on their difficulty and feature dissimilarity. Our method consistently reduces ECE and generally improves accuracy (yielding lower error rates), especially in low-label settings. This enhances the reliability of AI systems in high-stakes applications while also lowering data annotation costs.
CalibrateMix contributes to positive societal impact, enabling safer deployments through better-calibrated predictions. For future work, we aim to extend CalibrateMix to other tasks such as object detection, semantic segmentation, and out-of-domain (OOD) scenarios.

\section*{Acknowledgement}
This research is supported in part by the NSF IIS award 2107518, a UIC Discovery Partners Institute (DPI) award, and a Google CAHSI award. Any opinions, findings, and conclusions expressed here are those of the authors and do not necessarily reflect the views of NSF, DPI, or Google CAHSI. We thank our anonymous reviewers for their constructive feedback, which helped improve the quality of our paper.

\bibliography{aaai2026}
\end{document}